\pdfoutput=1
\documentclass[11pt]{article}

\usepackage{acl}

\usepackage{times}
\usepackage{latexsym}

\usepackage{times}
\usepackage{soul}
\usepackage{natbib} 
\usepackage{url}
\usepackage{xcolor}
\usepackage{graphicx}
\usepackage{amsmath}
\usepackage{amsthm}
\usepackage{subcaption}
\usepackage{booktabs}
\usepackage{algorithm}
\usepackage{algorithm}
\usepackage{algpseudocode}
\usepackage{amsmath}
\usepackage{cleveref}
\usepackage{amssymb}

\algblock{Input}{EndInput}
\algnotext{EndInput}
\algblock{Output}{EndOutput}
\algnotext{EndOutput}

\DeclareMathOperator*{\argmax}{arg\,max}

\Crefname{equation}{Eq.}{Eqns.}
\Crefname{figure}{Fig.}{Figs.}
\creflabelformat{equation}{(#2#1#3)}

\usepackage[T1]{fontenc}
\usepackage[utf8]{inputenc}

\usepackage{microtype}

\title{Automatic Rule Induction for Efficient Semi-Supervised Learning}

\author{
Reid Pryzant, Ziyi Yang,  Yichong Xu, Chenguang Zhu, Michael Zeng \\
Microsoft Cognitive Services Research Group\\
\{reidpryzant,ziyiyang,yicxu,chezhu,nzeng\}@microsoft.com
}

\begin{document}
\maketitle
\begin{abstract}
Semi-supervised learning has shown promise in allowing NLP models to generalize from small amounts of labeled data. Meanwhile, pretrained transformer models act as black-box correlation engines that are difficult to explain and sometimes behave unreliably. In this paper, we propose tackling both of these challenges via Automatic Rule Induction (ARI), a simple and general-purpose framework for the automatic discovery and integration of symbolic rules into pretrained transformer models. First, we extract weak symbolic rules from low-capacity machine learning models trained on small amounts of labeled data. Next, we use an attention mechanism to integrate these rules into high-capacity pretrained transformer models. Last, the rule-augmented system becomes part of a self-training framework to boost supervision signal on unlabeled data. These steps can be layered beneath a variety of existing weak supervision and semi-supervised NLP algorithms in order to improve performance and interpretability. Experiments across nine sequence classification and relation extraction tasks suggest that ARI can improve state-of-the-art methods with no manual effort and minimal computational overhead. 
\end{abstract}

\section{Introduction}

Large-scale pretrained neural networks can struggle to generalize from small amounts of labeled data \citep{devlin2019bert}, motivating approaches that leverage both labeled and unlabeled data. This is partially due to the black-box and correlational nature of neural networks, which confers the additional difficulties of uninterpretability \cite{bolukbasi2021interpretability} and unreliability \citep{sagawa2020investigation}.

A growing body of research seeks to ameliorate these issues by augmenting neural networks with symbolic components: heuristics, logical formulas, program traces, network templating, blacklists, etc \cite{arabshahi2018combining,galassi2020neural,wang2021neural}. In this paper, we refer to these components as \textit{rules}. Symbolic reasoning has attractive properties. Rules need little or no data to systematically generalize, and rules are inherently interpretable with respect to their constituent operations.

In this paper we propose a general-purpose framework for the automatic discovery and integration of symbolic rules into pretrained models. The framework contrasts with prior neuro-symbolic NLP research in two ways. First, we present a fully automatic rule generation procedure, whereas prior work has largely focused on manually crafted rules \cite{mekala2020contextualized,awasthi2020learning,li2021bertifying} or semi-manual rule generation procedures  \cite{boecking2020interactive,galhotra2021adaptive,zhang2022prboost}. With these existing techniques, practioners must formulate and implement their rules by hand, creating a second-order ``rule annotation'' burden on top of the data labeling process. 
 
Second, the proposed framework is general purpose and can be applied to any classification dataset. This contrasts with prior research that proposes task- and domain-specific symbolic logic, through weak supervision signals \citep{ratner2017snorkel,awasthi2020learning,safranchik2020weakly}, special loss functions \citep{xu2018semantic}, model architectures \citep{seo2021controlling}, and prompt templates \cite{schick2020exploiting}.

Our framework consists of two steps. First, we generate symbolic rules from data. This involves training low-capacity machine learning models on a reduced feature space, extracting artifacts from these models which are predictive of the class labels, then converting these artifacts into rules. After the rule induction step, we use the induced rules to amplify training signal in the unlabeled data. In particular, we adopt a rule-augmented self-training procedure, using an attention mechanism to aggregate the predictions of a backbone classifier (e.g. BERT) and the rules.

We evaluate the ARI framework across nine text classification and relation extraction tasks. The results suggest that the proposed algorithm can exceed state-of-the-art semi-supervised baselines, and that these gains may be because the model learns to rely more heavily on rules for difficult-to-predict examples. We also show that the proposed rule induction strategy can rival human crafted rules in terms of their quality. Last, we demonstrate the interpretabiltiy of the overall system. In summary, the contributions of this paper are:\footnote{An open-source implementation of the framework is available at: \url{https://github.com/microsoft/automatic-rule-induction}.}
\begin{itemize}
    \item Methods for automatically inducing and filtering symbolic rules from data.
    \item A self-training algorithm and attention mechanism for incorporating these rules into pretrained NLP models.
    \item Evidence suggesting the proposed framework can be layered beneath a number existing algorithms to boost performance and interpretability.
\end{itemize}

\section{The ARI Framework}

\begin{figure}
\centering
\includegraphics[width=\linewidth]{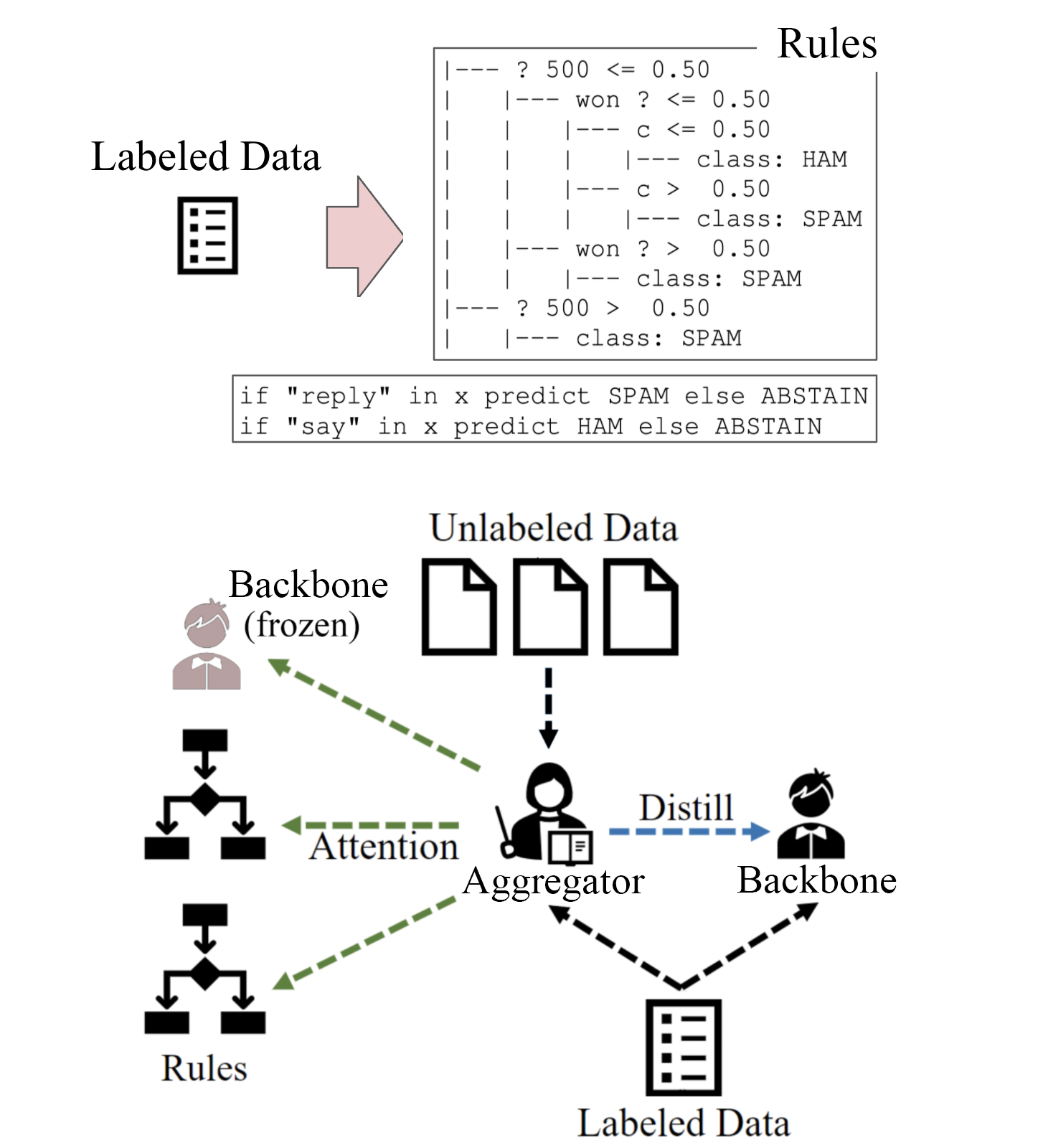}
\caption{Overview of the proposed Automatic Rule Induction (ARI) framework. First, rules are induced from labeled data (top, shown with real example rules). Second, the rules are integrated into pre-trained NLP models via an attention mechanism and a self-training procedure (bottom).}
\label{fig:model}
\end{figure}

The proposed rule induction framework seeks to automatically induce symbolic rules from labeled data. Next, the rules can be used to amplify training signal on the unlabeled data. These steps are depicted in \Cref{fig:model}.

More formally, assume we are given a target classification task consisting of labeled classification data $\mathcal{L} = \{(x_i, y_i)\}_{i=1}^M$ and unlabeled data $\mathcal{U} = \{(x_{i+M})\}_{i=1}^{N}$,
% \cz{rename x here as xi has already been used}, 
where each $x_i$ is a text string and $y_i \in \{1, ..., K\}$. Our proposed method uses the labeled data $\mathcal{L}$ to generate a set of symbolic prediction functions (``rules'') $\mathcal{R} = \{r_j\}_{j=1}^R$ that take the text and output a label or abstain: $r_j(x) \in \{-1\} \cup \{1, ..., K\}$. We then train a joint system which models $P(y \vert x;\ \mathcal{L}, \mathcal{U}, \mathcal{R})$, i.e., an estimator which utilizes the labeled data, unlabeled data, and rules to make reliable and interpretable predictions.

\subsection{Rule Induction}
We begin by explaining our rule induction technique. Concretely, the goal is to generate a set of prediction functions which use the text to output a label or abstain. We operationalize this as a three-stage pipeline. First, we featurize the text. Second, we use these features to construct rule-based predictor functions. Last, we filter the rules in order to block them from firing on risky examples (to maximise precision). %The output of this process is a set of rules which can be applied to unlabeled data in order to amplify training signal. 

\textbf{Text Featurization.} In the first step, the input text $x_j$ is converted into a binary or continuous feature space $\phi(x_j) \in \mathbb{R}^d$ that is more amenable to symbolic reasoning than the raw text.

\begin{enumerate}
    \item \emph{Ngram} ($\phi^N$). We adopt a bag-of-words model of the text, converting each string into a binary vector reflecting the presence or absence of words in a vocabulary of size $V$. 
    \item \emph{PCA} ($\phi^P$). Intuitively, if we only have a small amount of labeled data, then common ngrams may be spuriously correlated with the labels. To tackle this issue, we follow \citet{arora2017simple,yang2021simple} by subtracting off a vector of shared information from each feature matrix. Specifically, we compute the first principal component $v$ of an ngram feature matrix $P \in \mathbb{R}^{(M+N)\times d}$ constructed from both labeled and unlabeled texts in a dataset, i.e., the $j$th row $P_{j, :} = \phi^N(x_j) : j \in [1, M+N]$. Then it follows that singular value decomposition~(SVD) of the ngram feature matrix is $P = U \Sigma V^{T}$. The first principal component $v$ is the most ``common'' part of all samples (e.g., common words), and is defined as the first column of $V \in \mathbb{R}^{d\times d}$. We then remove the projection of all features vectors $\{\phi^N(x)\}$ onto $v$:
    \[
    \phi^P(x) := \phi^N(x) - v\frac{v^T \phi^N(x)}{\|\phi^N(x)\|_2} \\
    \]
    
    We hypothesize that this can help remove common information that is shared across many texts, in order to isolate the most unique and salient lexical phenomena.
    % \item \emph{Neuron-binary}. We save the neuronal activations of a pretrained BERT model, then binarize each neuron's activity according to whether it's activation is above or below it's average activation across the training and unlabeled datasets. 
    % \item \emph{Neuron}. We save the neuronal activations of a pretrained BERT model, then standardize each neuron's activity to have mean of 0 and standard deviation of 1.
\end{enumerate}

\textbf{Rule Discovery.}
Armed with a featurization of the texts in $\mathcal{L}$, we proceed by generating symbolic rules from the features which are capable of predicting the labels with high precision. In practice, these rules are artifacts of low-capacity machine learning models. We experiment with two rule generation algorithms.

The first rule generation algorithm uses a \emph{linear} model and can be applied to ngram-based (binary) feature spaces. It involves training a simple linear model $m(x_j) = \sigma(\mathbf{W} \phi(x_j))$ containing one matrix of parameters $\mathbf{W} \in \mathcal{R}^{K \times V}$ that predicts class labels from the input features. It is trained by using a cross-entropy loss function and $l_2$ regularization term \citep{tibshirani1996regression}. Note that in this case $\sigma$ represents an element-wise sigmoid function \citep{mao2014deep}. Next, we select the $R$ largest weights in $\mathbf{W}$ and create one rule from each weight. If a selected weight $w_{i, k}$ corresponds to feature $f_i$ and label $k$, then we create a rule $r$ that predicts label $k$ if the $i^{th}$ dimension of $\phi(x_j)$ is 1, otherwise abstaining:
\[   r(x_j) =  \begin{cases}
	k & \text{if }\phi_i(x_j)=1\\
	-1 & \text{otherwise}
	\end{cases}
\]

The second rule generation algorithm uses decision trees and can be applied to ngram- or PCA-based (binary or continuous) feature spaces. Intuitively, we want to find regions inside the range of each feature (or combination of features) that are predictive of the labels. We accomplish this by training a random forest classifier containing $R$ decision trees at a depth of $D$ (we use $D=3$ in the experiments). To make a rule from each decision tree, we apply a confidence threshold $\tau$ to the predicted label distribution in order to control the boundary between prediction and abstainment. In other words, if a decision tree $t_i$ outputs a probability distribution $\hat{p}$ over the labels, i.e. $t_i(\phi(x_j)) = \hat{p}_{i,j}$ then we construct a rule $r_i$ such that:
\[ r_i(x_j) = \begin{cases}
	\argmax( \hat{p}_{i,j})  & \text{if }\max( \hat{p}_{i,j})  > \tau \\
	-1 & \text{otherwise}
	\end{cases}
\]

 Note that due to the bagged construction of the random forest, we hypothesize that these decision trees will yield rules which can be aggregated for robust supervision signal. %Furthermore,  since each decision tree is operating in a symbolic fashion, the rule is interpretable and can be edited. 

\textbf{Rule Filtering.} Since rules are allowed to abstain from making predictions, we can introduce dynamic filtering mechanisms that block rules from firing on examples where the rule is likely to make errors. This helps increase the precision of our rules and increase the fidelity of our downstream rule integration activities.

\begin{itemize}
    \item \emph{Training accuracy.} The rules are not perfect predictors and can make errors on the training set. We randomly sample a proportion of these errors (50\% in the experiments) and replace the incorrectly predicted value with abstainment (-1).
    % \item \emph{Validation performance.} We compare the relative performance of each rule on a held-out validation set of labeled examples using F1 score. In order to compute each rule's validation score we only consider the subset of examples where the rule did not abstain. We then remove the bottom $x$ percent of rules ($16\%$ in the experiments) according to this score.
    \item \emph{Semantic coverage.} We design a filter to ensure that the ``covered'' subset of examples (examples where at least one rule fires) resembles the training set. In detail, after a rule $r_i$ fires on input text $x_j$, predicting label $r_i(x_j) = l$, we use the Sentence BERT framework \citep{reimers2019sentence} and a pre-trained mpnet model \citep{song2020mpnet} to obtain embeddings for the input sentence $x_j$ and all training samples that have the same label as the rule's prediction: $\{x_i \in \mathcal{L}\ :\ y_i = l\}$. We then compute the cosine similarity between the input's embedding and the training set embeddings. If the maximum of these similarities is below some threshold (0.8 in the experiments) then we block the rule $r_i$ from firing and replace its prediction $l$ with abstainment (-1).\footnote{Note that in applied settings, this may be computed on the fly with fast similarity search packages, e.g. \citet{johnson2019billion}. For this initial work, we pre-computed all filters prior to model training.}
\end{itemize}

\subsection{Rule Integration}
\label{sec:rule-integration}
After we have induced weak symbolic rules $\{r_i\}_{i=1}^R$ from the labeled data $\mathcal{L}$, we can leverage the rules and unlabeled data $\mathcal{U}$ for extra training signal. 

Our method is inspired by recent work in weak supervision and semi-supervised learning \cite{karamanolakis2021self,du2020self}. It consists of a backbone classification model (e.g. BERT) and a proposed rule aggregation layer. The aggregation layer uses an attention mechanism to combine the outputs of the backbone model and rules. The parameters of the backbone and aggregator are jointly trained via a self-training procedure over the labeled and unlabeled data. 

%We adopt the terminology of student/teacher training, where the ``student'' is a base classifier and the ``teacher'' is a trainable aggregation function that combines the predictions of the student and rules. Note that this usage the knowledge distillation literature, where the student and teacher are different models \cite{gou2021knowledge}.

In more detail, the \textbf{backbone} model $b(\cdot)$ is a standard BERT-based classifier with a prediction head attached to the \texttt{[CLS]} embedding. This classifier outputs a probability distribution over the possible labels. 

The \textbf{aggregation} layer $a(\cdot)$ is trained to optimally combine the predictions of the backbone model and rules. It does so via the following attention mechanism. The layer first initializes trainable embeddings $e_j$ for each rule $r_j$, and embedding $e_s$ for the backbone. Next, it computes dot-product attention scores between these embeddings and an embedded version of the input text ($h_i$). The final model prediction is a weighted sum of the backbone and rule predictions, where the weights are determined by the attention scores. 

Specifically, if the set of rules activated on input $x_i$ is $R_i = \{r_j \in \mathcal{R} : r_j(x_i) \neq -1\}$, and the function $g(\cdot)\in \mathcal{R}^K$ returns a one-hot encoding of its input, then the rule aggregation layer computes a probability distribution over the labels:
\begin{align}
a(x_i) = \frac{1}{Q} \left( \sum_{j: r_j \in R_i} s^j_i\ g(r_j(x_i)) + s^s_i\ b(x_i)  + u \right)
\end{align}
where the attention scores are calculated as,
\[
s^j_i = \sigma(p(h_i) \cdot e_j) \\
\]
Note that $p$ is a multi-layer perceptron that projects the input representation $h_i$ into a shared embedding space, $Q$ is a normalizing factor to ensure $a(x_i)$ is a probability distribution, $\sigma(\cdot)$ is the sigmoid function. Following \citet{karamanolakis2021self}, the quantity $u$ is a uniform smoothing term.

In order to train the overall system, we first pretrain the backbone on the labeled data $\mathcal{L}$. Next we iteratively co-train the backbone and aggregation layer. We train the aggregator (freezing the parameters of the backbone), then train the backbone (freezing the aggregator). The process is as follows:

\begin{table*}[]
\begin{center}
\begin{tabular}{@{}llllllllll@{}}
\toprule
             & AGNews & CDR  & ChemProt & IMDB  & SciCite & SemEval & SMS  & TREC & Youtube \\ \midrule
Domain      & News & Bio & Bio & Review & CS & Web & Sms & Speech & Web\\
\# Labeled   & 4800   & 421  & 643      & 1000  & 412     & 87      & 228  & 248  & 79      \\
\# Unlabeled & 91200  & 8009 & 12218    & 19000 & 7831    & 1662    & 4343 & 4717 & 1507    \\
\# Valid     & 1500  & 920  & 1500     & 1500  & 916     & 178     & 500  & 500  & 120     \\
\# Test      & 12000  & 4673 & 1607     & 2500  & 1861    & 600     & 500  & 500  & 250     \\
\# Classes   & 4      & 2    & 10       & 2     & 3       & 9       & 2    & 6    & 2       \\
\bottomrule
\end{tabular}
\caption{Datasets used in our experiments.}
\label{table:dataset}
\end{center}
\end{table*}

\begin{enumerate}
    \item Train the backbone $s$ using labeled data $\mathcal{L}$ and a cross-entropy loss function, where $b(x_i)_{y_i}$ denotes the logit for the groundtruth class $y_i$:
    \[\mathcal{\ell}_{stu}^{sup} = - \sum_{(x_i, y_i) \in \mathcal{L}} \log b(x_i)_{y_i} \]
    \item Repeat until convergence:
    \begin{enumerate}
        \item Train the aggregator $t$ on labeled data using a cross-entropy loss function :
\[\mathcal{\ell}_{tea}^{sup} = - \sum_{(x_i, y_i) \in \mathcal{L}} \log a(x_i)_{y_i} \]
        \item Train the aggregator on unlabeled data $\mathcal{U}$ with a minimum entropy objective \citep{grandvalet2004semi}. This encourages the aggregator to learn attention scores that favor rule agreement, because the aggregator will be encouraged to output more focused probability distributions, thereby placing less importance on spurious rules that disagree:
\[\mathcal{\ell}_{tea}^{unsup} = - \sum_{x_i \in \mathcal{U}} a(x_i)^T \log a(x_i) \]
where $\log a(x_i) \in \mathbb{R}^K$ denotes the element-wise logarithm of the probability distribution $a(x_i)$.
        \item Train the backbone on labeled data using $\ell_{stu}^{sup}$:
    \[\mathcal{\ell}_{stu}^{sup} = - \sum_{(x_i, y_i) \in \mathcal{L}} \log b(x_i)_{y_i} \]
        \item Train the backbone on unlabeled data by distilling from the aggregator, i.e. train the backbone to mimic the aggregator's output:
        \[
\mathcal{\ell}_{stu}^{unsup} =  - \sum_{x_i \in \mathcal{U}} a(x_i)^T \log b(x_i)
\]
    \end{enumerate}
\end{enumerate}

Once trained, one can use the outputs of either the backbone or aggregator for inference. If one uses the aggregator, they receive the benefit of improved interpretability: one could inspect the attention scores $s_i^j$ to understand what proportion of the system's decision was due to each rule.\footnote{Recent research shows that attention distributions in hidden layers are not valid explanations \citep{wiegreffe2019attention}, however in our case the attention scores are part of the model's output layer, i.e. used in a linear combination to calculate output probabilities directly. See \Cref{sec:analysis} for details.}

\section{Experiments}

We perform experiments across 9 datasets and tasks, finding that the ARI rule induction framework can improve the performance of state-of-the-art semi-supervised text classification algorithms. Note that concrete examples of the human-readable rules succeeding (and failing) are given in the Appendix.  

\subsection{Experimental Setup}
\label{sec:setup}
We evaluate our framework on nine benchmark NLP classification datasets that are popular in the few-shot learning and weak supervision literature \cite{ratner2017snorkel,awasthi2020learning,zhang2021wrench,cohan2019structural}. These tasks are as follows: \textbf{AGNews}: using news headlines to predict article topic, \textbf{CDR}: using scientific paper excerpts to predict whether drugs induce diseases, \textbf{ChemProt}: using paper experts to predict the functional relationship between chemicals and proteins, \textbf{IMDB}: movie review sentiment, \textbf{SciCite}: classifying citation intent in Computer Science papers, \textbf{SemEval}: relation classification from web text, \textbf{SMS}: text message spam detection, \textbf{TREC}: conversational question intent classification, \textbf{Youtube}: internet comment spam detection.

\Cref{table:dataset} shows dataset statistics. Our benchmarks cover a range of discourse domains and classification types. Unless otherwise stated we consider a 5\% / 95\% split between labeled data and unlabeled data. We construct this split by randomly partitioning the total training data and removing labels from the 95\% split. Following \citet{gao2020making,zhang2022prboost} we subsample each validation set so that it roughly matches the size of the training set in order to better simulate label scarcity.

% \subsection{Configuration}

All reported results are the average of ten experimental trials, each with different random splits, seeds, and initializations. For each trial, we continuously train our models for 12,500 steps using a batch size of 32, and we stop the training process early based on validation set performance. For each method (baseline and proposed), we conducted a minimal hyperparameter search (details in the Appendix) to establish the best validation performance before running inference over the test set. We ran all experiments on Microsoft Azure cloud compute using NVIDIA V100 GPUs (32G VRAM). All algorithms were implemented using the Pytorch and Wrench frameworks \cite{paszke2017automatic,zhang2021wrench}. We report binary F1 score for binary classification tasks and macro-weighted F1 for multiclass classification tasks. 

 %For the proposed automatic rule induction methods, we generated $R$ = 16, 32, or 64 rules based on validation performance and and ran 25 iterations of self-training. See the Appendix for more experimental details.

% \usepackage{booktabs}
\begin{table*}[tb!]
\centering
\small
\begin{tabular}{@{}lcccccccccc@{}}
\toprule
 Methods           & AGNews & CDR   & ChemProt & IMDB  & SciCite & SemEval & SMS   & TREC  & Youtube & Avg. \\ \midrule
\multicolumn{2}{l}{\textit{Baselines}}           &       &          &       &         &         &       &       &         &   \\
\ \ BERT              & 90.61  & 54.92 & 58.46    & 87.46 & 81.88   & 59.76   & 95.13 & 83.38 & 93.28   &   76.72   \\
\ \ Weak Ensemble    & 83.76  & 45.12 & 41.98  &  83.46  &  63.97 & 48.23 & 79.33 & 61.68 & 88.65 & 66.24 \\
\ \ LMFT              & 90.59  & 54.19 & 58.38    & 87.48 & 82.21   & 60.68   & 95.59 & 86.52 & 93.45   &  77.13    \\
\ \ Self-Train        & 91.30       & 55.58 & 54.90     &   88.76    & 81.15   & \textbf{68.1}    & 94.99 & 87.07 & 93.41   &  77.89    \\
\ \ Snuba & 90.46 &	53.99 &	58.27 &	87.29 &	82.03 &	60.57 &	95.43 &	86.35 &	93.30 &	76.98 \\

\ \ Min Entropy    & 90.97  & 55.14 & 56.00  &  89.05  &  82.01 & 63.70 & 95.03 & 84.92 & 93.14 & 77.27 \\
\ \ MoE & 89.94 & 54.44 & 57.27 & 87.25 & 81.83 & 60.93 & 95.03 & 85.24 & 92.53  &  76.65 \\ 
\ \ VAT & 91.31 & 56.01 & 55.70 & 88.49 & 81.12 & \textit{67.99} & 95.56 & \textbf{88.27} & 93.19 & 78.19  \\
\ \ PET               & \textbf{91.46}  &    51.16    & 53.90    & 88.49 &   75.13      &   64.45      & \textit{95.68} & 83.42 & \textbf{95.36}   &   75.76   \\
% LM-BFF            &        &       &          &       &         &         &       &       &         &      \\ \midrule643 + 
\midrule
% Snork + ngram     &        &       &          &       &         &         &       &       &         &      \\
% Snork + ngram-pca &        &       &          &       &         &         &       &       &         &      \\
\multicolumn{2}{l}{\textit{ARI (proposed)}}           &       &          &       &         &         &       &       &         &   \\
\ \ Ngram + linear       &  \textit{91.37}      & 56.11 & \textit{60.08}    &  \textit{89.10}     & 81.51   & 65.56   & 95.6  & 85.43 & \textit{95.24}   &  78.56    \\
\ \ Ngram + tree      &  91.11     &  \textit{57.77}     &  \textbf{60.95}        &    \textbf{89.41}   &  \textit{82.76}       &  64.01       &   92.87    &  86.99     &  93.55        &  \textit{78.62}   \\
\ \ PCA + tree   & 90.87   & \textbf{57.92} & 60.01    &    88.33   & \textbf{83.76}   & 65.10    & \textbf{95.74} & \textit{87.19} & 94.95   &   \textbf{79.05}   \\  \midrule
\multicolumn{2}{l}{\textit{Oracles}}           &       &          &       &         &         &       &       &         &   \\ 
\ \ ASTRA   &    91.71    &  61.63  &    59.58      &   88.98    &    82.29     &     75.18    &   93.15    &  87.23  &   96.42      &   80.70\\
\ \ T-few   &    95.12    &  57.23  &    55.84      &   94.48    &    84.21     &     64.9    &   96.73    &  89.13  &   96.84      &   79.79 \\
\bottomrule
\end{tabular}
\caption{Semi-supervised learning performance on nine classification datasets. Following \citet{tatiana2021not}, we report the geometric mean in the ``Avg.'' column. We denote the highest and second-highest performance (excluding the expert rules model) in \textbf{bold} and \textit{italic} respectively. Note that T-Few was pretrained on AGNews, TREC, and IMDB. See \Cref{tab:acc} for accuracies and comparison against the PR-BOOST algorithm \cite{zhang2022prboost}. \label{table:performance_main} }
\end{table*}

\subsection{Baselines}

We experiment with our \textbf{ngram} and \textbf{pca}-style featurization schemes, as well as our linear model (\textbf{linear}) and decision tree (\textbf{tree})-based rule generation methods. We compare against the following baselines:

% \begin{itemize}
\noindent \textbf{BERT}: directly fine-tuning a BERT model on the available supervised data \cite{devlin2019bert}. 
\noindent \textbf{Weak Ensemble}: It is possible that traditional ML models like regressions and decision trees achieve good performance in these low-resource settings, and the proposed ARI framework just takes advantage of these models. We accordingly train several weak models (BERT, regression, and random forest using the same hyperparameters as was used to obtain rules) and ensemble their predictions for comparison. 
\noindent \textbf{LMFT}: training a BERT model on the unlabeled data with its original language modeling objective before fine-tuning on the supervised data \cite{howard2018universal,gururangan2020don}. 
\noindent \textbf{Self-Train}: iteratively self-training towards the predictions of a frozen model on the unlabeled data \cite{nigam2000analyzing,lee2013pseudo}.
\noindent \textbf{Snuba}: We use the Snuba algorithm \cite{varma2018snuba} to automatically generate weak labels over the unlabeled data, then a generative label model from Snorkel \cite{ratner2017snorkel} to expand the available training data prior to BERT fine-tuning. This baseline offers a direct comparison against a popular weak supervision procedure.
\noindent \textbf{Min Entropy}: Multitask self-training with a minimum entropy objective on the unlabeled data \cite{grandvalet2004semi}.
\noindent \textbf{MoE}: This is the same as the model proposed in \Cref{sec:rule-integration} except the rules are replaced with two-layer neural network classifiers that are trained end-to-end with the rest of the system. This baseline tests whether the proposed training procedure has the potential of achieving higher accuracy without the rule induction step. This baseline is similar to having a Mixture of Experts layer at the output \cite{jacobs1991adaptive,shazeer2017outrageously} without input routing, expert gating, or load balancing. 
\noindent \textbf{VAT}: Multitask self-training with a virtual adversarial regularization penalty on   the unlabeled data \cite{miyato2018virtual}. 
\noindent \textbf{PET}: a state-of-the-art method for semi-supervised learning that leverages prompting and model ensembling \cite{schick2020exploiting}. Note that PET is not a fully automatic procedure as it requires prompt templates and class verbilizations for each dataset. We used domain intuition to verbalize each class label, and constructed two prompt templates for each task: ``\texttt{[MASK] : [example]}'' and ``\texttt{[MASK] : [domain word]: [example]}'' where \texttt{[domain word]} is a word that signifies the nature of the ensuing text (e.g. ``\texttt{Review}'' for the IMDB dataset). See Appendix for details.
    % \item \textbf{LM-BFF}: a state-of-the-art method for few-shot learning \cite{gao2020making} which (1) automatically generates prompts and class label words for sentence completion style fine-tuning, and (2) expands the input context with demonstration examples from the training set.

We also compare against two oracles. The first called \textbf{ASTRA} and is a state-of-the-art weak supervision algorithm that uses manually designed rules and an iterative self-training procedure  \cite{karamanolakis2021self}. For this oracle we use previously published heuristic labeling functions from the weak supervision literature \cite{zhang2021wrench}. The rules were manually constructed using domain expertise and, being expertly crafted, suggest an upper bound on performance. The second oracle is called \textbf{T-Few} \cite{liu2022few} and represents a state-of-the-art prompting approach using a large 3 billion parameter model (30 times larger than the rest of the models considered in this paper).

\subsection{Experiment Results}
\label{sec:results}
% Our results show that ARI outperforms the baselines on most datasets, and its performance is also on par with the ASTRA model based on expert rules. Somewhat surprisingly, ARI outperforms ASTRA on four of the nine datasets that we experimented with.

\paragraph{Overall results.} \Cref{table:performance_main} presents our main results. The proposed ARI framework achieves the best performance on 5 out of 9 datasets, and the ARI variations beat the baselines in terms of average performance. Our results suggest that LMFT does not always improve the performance over standard BERT finetuning, and can hurt the performance sometimes (CDR). This is in line with previous research findings \citep{vu2021strata,du2020self}. Self-Train achieves an overall better performance than BERT, but underperformed on ChemProt and overperformed on SemEval. PET achieves strong results on AGNews and Youtube, but fails on many other datasets. This might be due to its sensitivity to prompts and label words for the scientific domains, which is typical for prompt-based models \citep{gao2020making}. Additionally, due to implementation differences in this prior work, we tested PET after a fixed number of training steps instead of the early-stopping validation technique employed by the other algorithms (\Cref{sec:setup}).

For ARI, decision-tree based methods give the best results overall, while there is no clear winner between PCA and Ngram-based models. Considering that we also removed stop words in the Ngram features, using PCA to remove common components might not make a big difference to the rules. The performance of ARI is close to ASTRA which uses manually crafted expert rules, showing the potential of automatic rules. Surprisingly, ARI is better than ASTRA on SciCite and SMS by a nontrivial margin. This suggests that automatic rules have the potential to rival human-generated rules. See the Appendix for further results and analysis.

Our results suggest that for prompting methods like PET and T-Few to outperform ARI, one needs bigger models with more language capacity like the 3B parameter Tfew (30x larger than e.g. BERT). ARI and PET leverage smaller models which are faster with reduced memory but also reduced capacity and therefore less effective prompts as prior research as noted \cite{liu2021pre}. In the Appendix, we observe that ARI continues to outperform PET when more powerful backbone encoders like DeBERTaV3 \cite{he2021debertav3} are used (\Cref{table:encoders}).

\paragraph{Robustness}

\begin{figure}[ht!]
	\centering
	\begin{subfigure}[b]{0.235\textwidth}
		%		\centering
		\includegraphics[width=\textwidth]{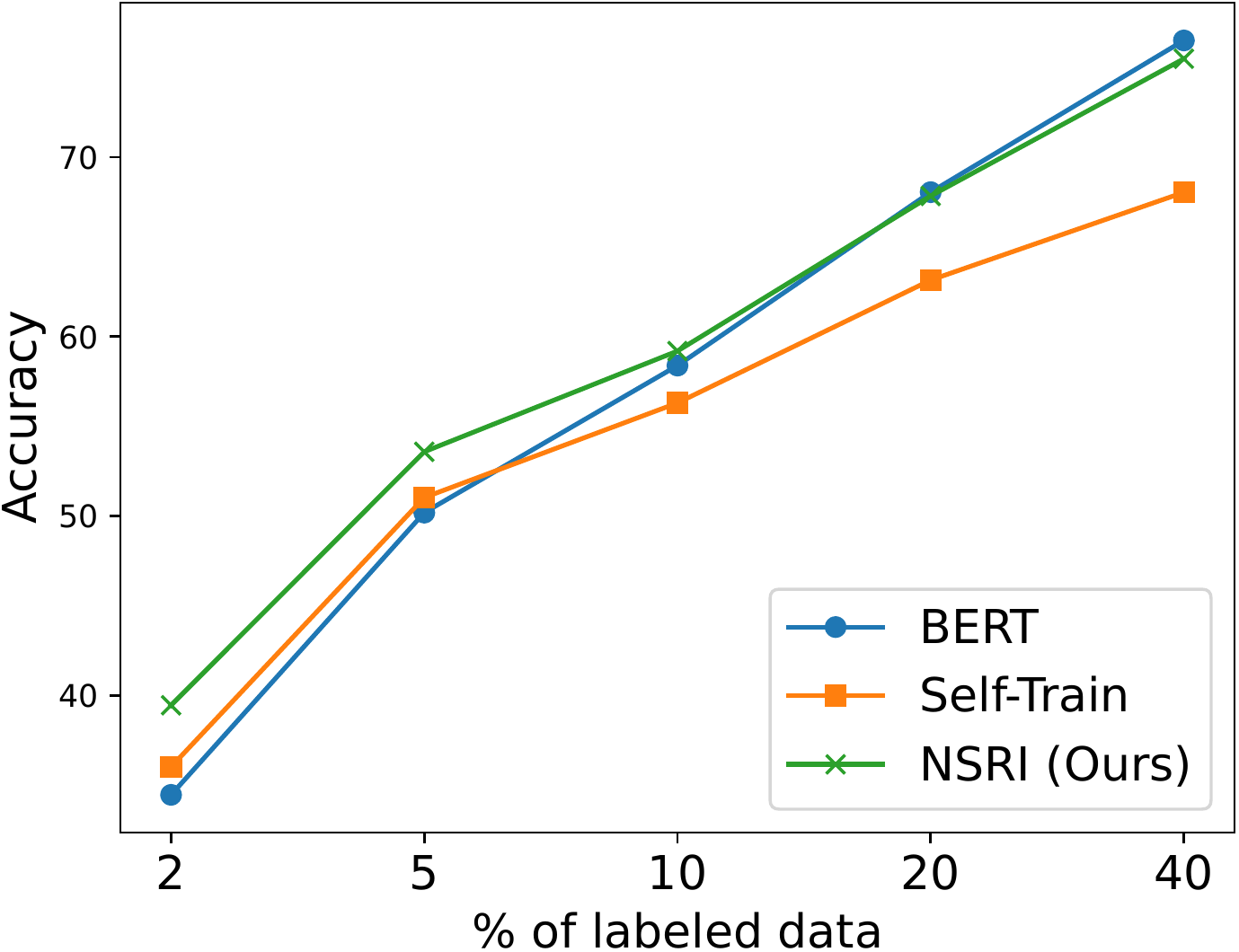}
		\caption{ChemProt}
	\end{subfigure}%
	\hspace{0.5mm}
	\begin{subfigure}[b]{0.235\textwidth}
		%		\centering
		\includegraphics[width=\textwidth]{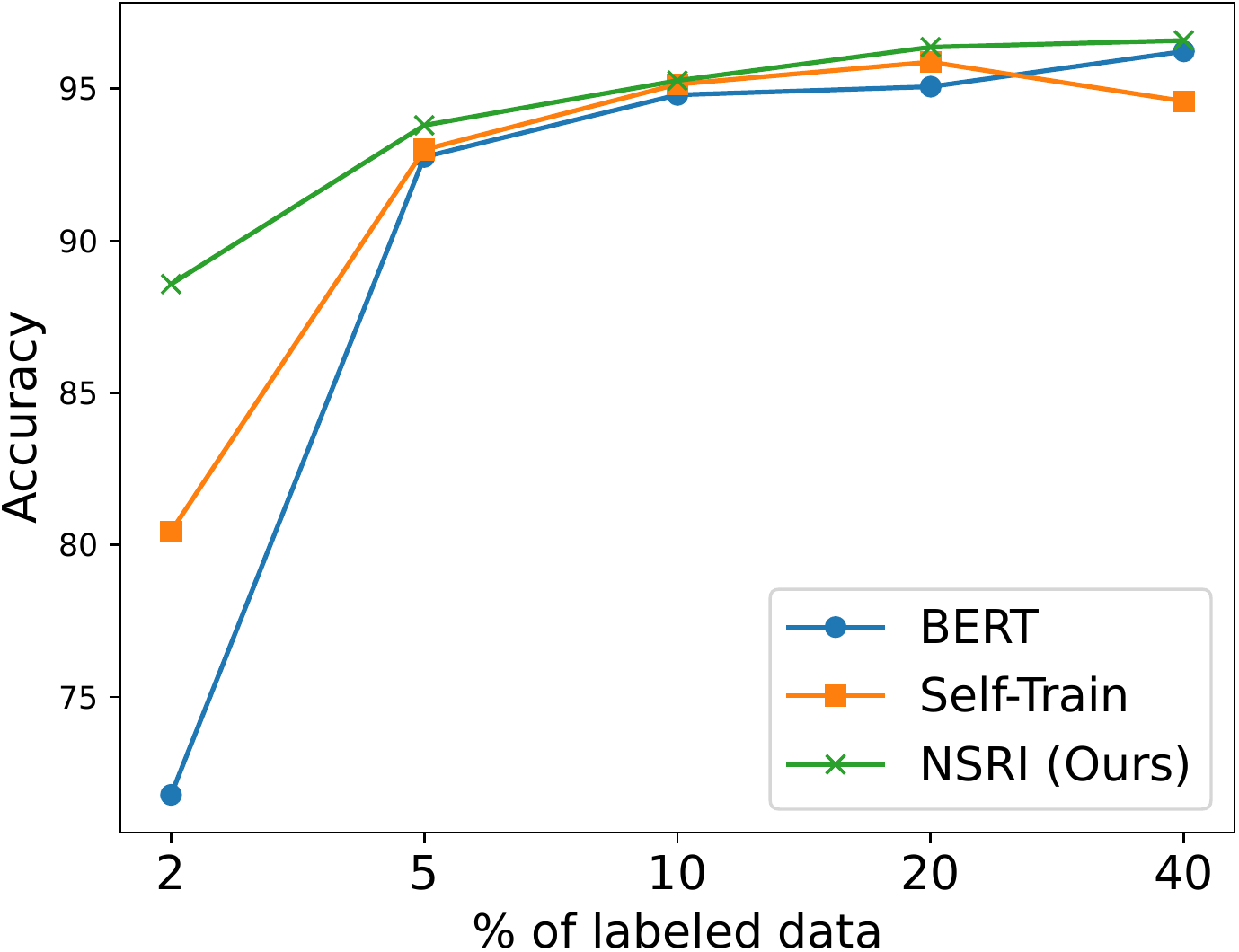}
		\caption{Youtube}
	\end{subfigure}%
	\caption{Robustness to training data size. 
	\label{fig:robustness}}
\end{figure}

We further test our method's robustness to the number of labeled examples in \Cref{fig:robustness}. We vary the fraction of labeled data between 2\% to 40\% on the ChemProt and Youtube datasets. The results suggest that ARI can reliably outperform the baselines across this range, especially when labeled data is scarce. Standard supervised BERT fine-tuning become increasingly competitive as the fraction of labeled data exceeds 40\%.

\paragraph{Filter Ablations. }
We provide ablation results on rule filtering methods in \Cref{table:ablation_filtering}. We pick the best performers between the three rule-generation methods in \Cref{table:performance_main} and then vary the filters. 
All the three methods show performance gains when applied individually, and combining the filters appears to further improve performance in some cases.

\begin{table}[H]
\small
\begin{tabular}{@{}lllll@{}}
\toprule
Filters                  & CDR   & SciCite & SemEval & SMS   \\ \midrule
None                        & 54.86 & 81.14   & 63.67   & 94.72 \\
%Valid                    & 55.13 & 81.82   & 64.16   & 94.77 \\
Semantic                 & \textbf{57.20} & 82.00   & 65.43   & 94.54 \\
Train                    & 56.84 & 81.05   & 64.36   & 95.21 \\
%Valid + Sem.         & \textbf{57.92} & \textbf{82.63}   & 65.38   & 94.99 \\
 Sem. + Train & 56.46 & \textbf{82.41}   & \textbf{65.56}   & \textbf{95.74} \\
\bottomrule
\end{tabular}
\caption{Ablation study on rule filtering methods. We show the best result between Ngram+linear, Ngram+tree and PCA+tree. \label{table:ablation_filtering}}
% filter ablations (picked the best performer from table to to experiment with}
\end{table}

\paragraph{Hard or Soft Labels?}
\label{sec:hardsoft}

There are many variations on the basic self-training algorithm. Some prior work only trains the student on a small number of unlabeled examples having the highest confidence \cite{rosenberg2005semi,mcclosky2006effective,sohn2020fixmatch}. Recently, other work has opted to train the student on all available data, regardless of teacher confidence  \cite{vu2021strata}. Additionally, self-training can be performed with either the whole distribution (soft labels) or max probability label (hard labels) \cite{karamanolakis2021self}.

Our results are in \Cref{table:selftrain} and suggest that while there is no clear winner between hard and soft labels, training the student on a broad distribution of data is best.

\begin{table}[H]
\small
\begin{tabular}{@{}lllll@{}}
\toprule
              & CDR   & SemEval & Youtube & Chemprot \\ \midrule
Hard          & 56.17 & 65.11   & \textbf{95.24}   & \textbf{60.95}    \\
Hard + Thresh  & 54.34 & 62.78   & 93.52   & 60.65    \\
Soft          & \textbf{57.92} & \textbf{65.56}   & 94.14   & 59.34    \\
Soft + Thresh & 55.6  & 62.77   & 94.09   & 57.88    \\ \bottomrule
\end{tabular}
\caption{
Variations on the self-training algorithm. 
\label{table:selftrain}}
\end{table}

\paragraph{PCA mean subtraction}
\label{sec:pcaabl}

It is common practice to center the data by subtracting the means of each covariate away from a feature matrix prior to computing PCA \cite{mishra2017multivariate}. Our ARI procedure did not employ this mean subtraction trick because intuitively, centering the data prior to PCA prevents the first PCs from being dominated by the variables’ means, but this “mean load” is what we intend to capture and control for with our procedure. To validate this hypothesis, we enabled mean subtraction for ARI on 3 datasets. \Cref{table:pca} gives the results and we can observe a slight decrease in performance, -0.32\% on average, when mean subtraction is enabled. 
% \usepackage{booktabs}
\begin{table}[]
\begin{tabular}{@{}llll@{}}
\toprule
                       & TREC  & SMS   & IMDB  \\ \midrule
ARI                    & 87.19 & 95.74 & 88.33 \\
ARI + mean subtraction & 86.71 & 95.6  & 87.97 \\ \bottomrule
\end{tabular}
\caption{PCA mean subtraction ablation (PCA + tree variant). \label{table:pca}}
\end{table}

\paragraph{Transferability of rules}
\label{sec:pcaabl}
One possible limitation of ARI is that the rules overfit to their data and are extremely limited to a specific setting. To investigate this, we experimented with swapping the rules between related spam detection datasets SMS and Youtube and found a slight drop in performance (SMS: 95.74 -> 95.26, Youtube: 94.95 -> 93.72). We also tried unrelated datasets IMDB and ChemProt and found that performance drops more (IMDB: 88.98 -> 88.1, ChemProt: 60.1 -> 55.89), but not severely, perhaps because the model learns to ignore most of the rules and defaults to plain self-training (rule attention scores support this: for the matched data, 48.3\% of attention went to rules on average while only 21.1\% for the unmatched data). We conclude that while the learned rules are adapted to their training data, they are indeed transferable to other domains to some degree, although we note that this work assumes access to in-domain training sets, and that out-of-domain or zero-shot generalization is outside the paper's scope.

\section{Interpretability}
\label{sec:analysis}

As discussed in \Cref{sec:rule-integration}, the behavior of the aggregation layer $a(\cdot)$ can be traced to individual rules, which are themselves human readable and interpretable. This is because the output of $a(\cdot)$ is a linear combination of attention scores and rule predictions (Equation 1). In other words, if the attention score for rule $r_j$ on example $x_i$ is $s^j_i$, then the strength of rule $r_j$'s contribution to the model's final prediction is exactly $s^j_i / Q$. 

See the Appendix for case studies showing the impact of individual rules on model behavior. 

To further demonstrate the system's interpretability, we grouped examples according to their difficulty\footnote{Following \cite{swayamdipta2020dataset}, we used the entropy of BERTs predicted label distribution as a measure of example difficulty. We ranked examples according to this measure, then split them into hard  (above the 75th percentile), medium (25-75th percentile) and easy (below 25th percentile).} and measured the cumulative effect of rules on model behavior (i.e., $\sum_j s^j_i/Q$) for each category. The results are given in \Cref{table:investigative}. We observe that much of ARI's gains come from the hard-to-predict examples, and that these difficult examples may be associated with increased rule reliance.

\begin{table}[H]
\small
\centering
\begin{tabular}{@{}llll@{}}
\toprule
           & Easy  & Medium & Hard \\ \midrule
\textit{Youtube} &      &   &     \\
Bert (Acc) & 100     & 95.96  & 79.36    \\
ARI (Acc) & 100 (\textit{45.2})     & 94.35 (\textit{48.5})  & 87.3 (\textit{49.2})     \\ \midrule
\textit{ChemProt} &      &   &     \\
Bert (Acc) & 94.27     & 77.83  & 43.78    \\
ARI (Acc) & 92.78 (\textit{57.4}) & 77.97 (\textit{62.9})  & 45.02 (\textit{62.4})    \\\bottomrule
\end{tabular}
\caption{
Model performance and rule reliance according to whether examples are easy, medium, or hard to predict. The average cumulative effect of the rules $\mathbb{E}_{i \in G} [\ \sum_j s^j_i/Q\ ]$ for each group $G$ is given in parentheses. 
\label{table:investigative}}
\end{table}

\section{Related Work}
% https://arxiv.org/pdf/2005.06133.pdf
%  https://www.vldb.org/pvldb/vol12/p223-varma.pdf
%  https://arxiv.org/pdf/2203.09735.pdf
Our research draws on a number of related areas of research, including Neuro-Symbolic computation, semi-supervised learning, and weak supervision. 

Neuro-symbolic approaches seek to unite Symbolic AI, which from the 1950's until the mid 1990's was the dominant paradigm of AI research \cite{crevier1993ai,russell2002artificial}, with statistical machine learning and neural networks. For example, there is work that uses discrete parses to template neural network components \cite{arabshahi2018combining,mao2019neuro,yi2018neural}. There is also work that seeks to embed symbolic knowledge into network parameters via special loss functions \cite{xu2018semantic,seo2021controlling} or carefully curated datasets \cite{lample2019deep,clark2020transformers,saeed2021rulebert} and architectures \cite{trask2018neural}.  Other related work seeks to incorporate logical constraints into text generation models \cite{wang2021neural,lu2020neurologic}.

Our framework is further inspired by semi-supervised learning research that leverages labeled and unsupervised data. Our baseline PET model comes from a family of algorithms that leverage prompting and model ensembling for greater data efficiency \cite{schick2020exploiting,schick2020s}. There is also research on pulling in demonstration examples from the training set \cite{gao2020making}, automatic prompt generation \cite{zhang2021differentiable,li2021prefix}, and leveraging extra datasets and tasks for data augmentation when data is scarce \cite{du2020self,vu2021strata}.

Our self-training approach is similar to the knowledge distillation literature \cite{hinton2015distilling,gou2021knowledge} where a ``student'' model is trained to imitate the predictions of a ``teacher'' model. In our case, the teacher is not a separate model but a frozen student plus rule aggregation layer. 

Another close body of research taps into weak sources of supervision like regular expressions, keywords, and knowledge base alignment \citep{mintz2009distant,augenstein2016stance,ratner2017snorkel}. Researchers have incorporated these weak supervision signals into self-training procedures like ours \cite{karamanolakis2021self}, as well as constructing procedural generators for boosting weak supervision signals \cite{zhang2021wrench} and interactive pipelines for machine-assisted rule construction \cite{zhang2022prboost,galhotra2021adaptive,maheshwari2020semi}. There is also research on automatically generating weak labeling functions \cite{varma2018snuba,maheshwari2021learning} which shares our bag-of-words featurization and regression scoring mechanism.

\section{Conclusion}

In this paper, we proposed Automatic Rule Induction (ARI), a simple and general-purpose framework for the automatic discovery and integration of symbolic rules into pretrained NLP models. Our results span nine sequence classification and relation extraction tasks and suggest that ARI can improve state-of-the-art algorithms with no manual effort and minimal computational overhead. 

Future work could investigate layering ARI beneath other few-shot and semi-supervised algorithms, and improving the underlying rule generation strategies, particularly with causal mechanisms \cite{feder2021causal}. % and introducing causal mechanisms for rule  and improved rule generation strategies. %   , e.g. by introducing causal filters, building rules with generative models and/or reinforcement learning. 

\section{Limitations}
ARI is not without limitations. We observe that hyperparameter selection is key for quality rule generation \cite{feurer2019hyperparameter}. Second, as other research has noted \cite{dodge2019show,xu2021fewclue}, few-shot evaluation protocols remain immature as they rely on small, high variance training sets and static test sets. Last, our procedure works by extrapolating correlations in small training sets, which may result in overfitting and undermine robustness to distribution shift \cite{sagawa2020investigation}. While the results in Table 2 suggest the end-to-end ARI system is not any more susceptible to spurious correlations than other ML/DL-based methods, i.e. susceptibility is not a big enough issue to prevent SOTA or near-SOTA performance. We hypothesize this may be due to two reasons. First, the rules are heavily regularized: our rule selection model has a strong l2 penalty, our decision trees are generated as part of a stochastic random forest, and the PCA subtraction may also have a regularizing effect. Second, ARI is a hybrid system (neural + rule) which can learn to favor the pre-trained student model when spurious rules fire. Our min-entropy loss function on unlabeled data is designed to encourage such behavior. Concrete examples of spurious rules being ignored can be found in Appendix E.  

\section{Ethical Considerations}  % TODO add enviro
% encouraged by paper call https://aclrollingreview.org/authorchecklist

Adding symbolic components to neural systems is a promising way to improve AI trust. Symbolic mechanisms are inherently more interpretable and controllable than black-box function approximators. These components can be reviewed by independent panels and modified to fit the considerations and sensitivities of particular applications. 

Microsoft has been 100\% carbon neutral since 2012, is committed to being carbon negative by 2030 and removing all of its historical emissions by 2050. This extends to the Microsoft Azure cloud compute engine used for our experiments, which runs on majority renewable energy \cite{cloudrenewable}. 

\section{Acknowledgements}
We thank Pengcheng He, Giannis Karamanolakis, Hannes Schulz, Yu Shi, Robert Gmyr, Yuwei Fang, Shuohang Wang and many others for their advice.

\bibliography{anthology,custom}
\bibliographystyle{acl_natbib}

\appendix

\section{Appendix A: Reproducibility}
\label{sec:appendixA}

To construct our ngram feature matrices, we built a vocabulary of size 1600 using NLTK's WordNet lemmatizer and \texttt{word\_tokenize} tokenizer. We used the built-in English stopwords list, as well as a max document frequency cutoff of 0.95 and minimum token frequency cutoff of 4, and ngrams up to length 2.

Hyperparameters are given below. For each algorithm we describe the search space and say in parentheses which settings had the best validation performance for each dataset (and thus were selected for testing). Unless otherwise stated, we used a learning rate of 1e-5 for all algorithms, a batch size of 24, max sequence length of 128, and optimized using Adam \cite{kingma2014adam}. Note that we used the originally published hyperparameters for the Min Entropy, VAT, and MoE baselines. 

\textbf{BERT}:
\begin{itemize}
    \item No search.
\end{itemize}

\textbf{LMFT}: 
\begin{itemize}
    \item Pretraining epochs: 1, 3 (all datasets), 5.
\end{itemize}

\textbf{Self-Train}: 
\begin{itemize}
    \item Number of self-training iterations: 15, 25 (all datasets), 40.
    \item Ratio of labeled-to-unlabeled train steps: 0.7 (all datasets), 1.0. 
\end{itemize}

\textbf{PET}: 
\begin{itemize}
    \item Learning rate 1e-6, 1e-5 (all datasets), 1e-4. 
    \item Ensemble model train epochs: 2 (AGNews, IMDB), 3 (CDR, ChemProt, SciCite, TREC), 5 (SemEval, SMS, Youtube).
    \item Final classifier train epochs: 2, 3 (everything else), 5 (SemEval, Youtube), 10.
    \item Our prompting templates are given in \Cref{table:pet}.
\end{itemize}  

\begin{table*}[]
\begin{tabular}{@{}llll@{}}
\toprule
         & Label words                                                                          & Template 1              & Template 2                       \\ \midrule
SMS      & normal, junk                                                                         & {[}mask{]} : {[}text{]} & {[}mask{]} message: {[}text{]}   \\
AGNews   & world, sports, business, tech                                                        & {[}mask{]}: {[}text{]}  & {[}mask{]} news: {[}text{]}      \\
IMDB     & bad, good                                                                            & {[}mask{]}: {[}text{]}  & {[}mask{]} review: {[}text{]}    \\
SemEval  & cause, component, content, destination, origin & {[}mask{]}: {[}text{]}  & {[}mask{]} text: {[}text{]}      \\
 &  instrument, member, message, product & & \\ 
CDR      & no, yes                                                                              & {[}mask{]}: {[}text{]}  & {[}mask{]} text: {[}text{]}      \\
SciCite  & background, method, result                                                           & {[}mask{]}: {[}text{]}  & {[}mask{]} text: {[}text{]}      \\
ChemProt & part, regulator, up, down, agony, antagonist         & {[}mask{]}: {[}text{]}  & {[}mask{]} Passage: {[}text{]}   \\
 &   modify, together, product, not & & \\ 
Youtube  & normal, junk                                                                         & {[}mask{]}: {[}text{]}  & {[}mask{]} comment: {[}text{]}   \\
TREC     & description, entity, human, abbreviation                           & {[}mask{]}: {[}text{]}  & {[}mask{]} statement: {[}text{]} \\
 &    location, number & & \\ \bottomrule
\end{tabular}
\caption{PET verbalization templates and label word mappings. \label{table:pet}}
\end{table*}

\textbf{ARI}:
\begin{itemize}
    \item Rule embedding size: 100. 
    \item Number of rules: 16 (AGNews, CDR, ChemProt, SMS, Youtube), 32 (IMDB, SciCite, SemEval, TREC), 64. 
    \item Inference with student (AGNews, CDR, IMDB, SciCite), teacher (ChemProt, SemEval, SMS, Youtube, TREC).
    \item Tree rule threshold: 0.95 (SciCite), 0.8 (all other datasets). 
    \item Number of self-training iterations: 15, 25 (all datasets), 40.
    \item Ratio of labeled-to-unlabeled train steps: 0.7 (all datasets), 1.0. 
    \item Filter selection: described in Section .
\end{itemize}

\section{Appendix B: Example Rules}

We provide some concrete examples of unigram rules generated by ARI on the SMS dataset. This dataset involves detecting whether text messages are spam or not, so they are relatively easy to reason about in an intuitive sense.  
\\ \\
\textbf{Ngram linear rules}:
{\small
\begin{verbatim}
if "..." in x predict HAM else ABSTAIN
if ": )" in x predict HAM else ABSTAIN
if ".." in x predict HAM else ABSTAIN
if "txt" in x predict SPAM else ABSTAIN
if "service" in x predict SPAM else ABSTAIN
if "." in x predict HAM else ABSTAIN
if "claim" in x predict SPAM else ABSTAIN
if "dating" in x predict SPAM else ABSTAIN
if "?" in x predict HAM else ABSTAIN
if "ringtone" in x predict SPAM else ABSTAIN
if "ok" in x predict HAM else ABSTAIN
if "reply" in x predict SPAM else ABSTAIN
if "say" in x predict HAM else ABSTAIN
if "free" in x predict SPAM else ABSTAIN
if "home" in x predict HAM else ABSTAIN
if "fancy" in x predict SPAM else ABSTAIN
\end{verbatim}}

Some of these rules make sense; text messages asking recipients to ``claim'' items that are ``free'' or ``fancy'' are probably spam. Smiley faces (``: )'') and proper punctuation (``.'', ``?'') are normal things to write in a text message.
\\ \\

\textbf{Ngram tree rules}:

Note that our random forest was implemented with the sklearn package \cite{trappenberg2019machine} and so we use the same display format as their  \texttt{sklearn.tree.export\_text}. function: each node evaluates the frequency of it's associated string and branches accordingly.

{\small
\begin{verbatim}
|--- mob week <= 0.50
|   |--- ? ? 1000 <= 0.50
|   |   |--- awarded ? <= 0.50
|   |   |   |--- class: HAM
|   |   |--- awarded ? >  0.50
|   |   |   |--- class: SPAM
|   |--- ? ? 1000 >  0.50
|   |   |--- class: SPAM
|--- mob week >  0.50
|   |--- class: SPAM


|--- ? 500 <= 0.50
|   |--- won ? <= 0.50
|   |   |--- c <= 0.50
|   |   |   |--- class: HAM
|   |   |--- c >  0.50
|   |   |   |--- class: SPAM
|   |--- won ? >  0.50
|   |   |--- class: SPAM
|--- ? 500 >  0.50
|   |--- class: SPAM


|--- urgent ! <= 0.50
|   |--- ringtone <= 0.50
|   |   |--- send stop <= 0.50
|   |   |   |--- class: HAM
|   |   |--- send stop >  0.50
|   |   |   |--- class: SPAM
|   |--- ringtone >  0.50
|   |   |--- class: SPAM
|--- urgent ! >  0.50
|   |--- class: SPAM


|--- ! <= 0.50
|   |--- 750 <= 0.50
|   |   |--- win <= 0.50
|   |   |   |--- class: HAM
|   |   |--- win >  0.50
|   |   |   |--- class: SPAM
|   |--- 750 >  0.50
|   |   |--- class: SPAM
|--- ! >  0.50
|   |--- . <= 0.50
|   |   |--- cash <= 0.50
|   |   |   |--- class: HAM
|   |   |--- cash >  0.50
|   |   |   |--- class: SPAM
|   |--- . >  0.50
|   |   |--- line <= 0.50
|   |   |   |--- class: HAM
|   |   |--- line >  0.50
|   |   |   |--- class: SPAM    
\end{verbatim}
}

We find that these rules are less readily interpretable than directly using ngrams, but generally make sense. For example, the second to last rule suggests that if a text message contains an exclamation mark and large number (750) followed by ``win'', the message is spam (``win 750\$!'') but without the word ``win'' the message is probably not spam, (there are plenty of non-spammy reasons to talk about large numbers in a text message). 

\textbf{PCA tree rules}:
These rules, being constructed from a dense feature space, are less readily interpretable. We denote each feature dimension by the ngram it originated from, wrapped in quotes and followed by (+PCA).

{\small \begin{verbatim}
    
|--- come (+PCA) <= -0.03
|   |--- '& free' (+PCA) <= -0.00
|   |   |--- class: HAM
|   |--- '& free' (+PCA) >  -0.00
|   |   |--- 'won' (+PCA) <= -0.02
|   |   |   |--- class: HAM
|   |   |--- 'won' (+PCA) >  -0.02
|   |   |   |--- class: SPAM
|--- 'come' (+PCA) >  -0.03
|   |--- 'ringtone' (+PCA) <= 0.50
|   |   |--- 'latest' (+PCA) <= 0.50
|   |   |   |--- class: HAM
|   |   |--- 'latest' (+PCA) >  0.50
|   |   |   |--- class: SPAM
|   |--- 'ringtone' (+PCA) >  0.50
|   |   |--- class: SPAM


|--- ''m' (+PCA) <= -0.05
|   |--- 'lt ; #' (+PCA) <= -0.13
|   |   |--- 'win' (+PCA) <= -0.02
|   |   |   |--- class: HAM
|   |   |--- 'win' (+PCA) >  -0.02
|   |   |   |--- class: SPAM
|   |--- 'lt ; #' (+PCA) >  -0.13
|   |   |--- 'win ? ?' (+PCA) <= -0.01
|   |   |   |--- class: HAM
|   |   |--- 'win ? ?' (+PCA) >  -0.01
|   |   |   |--- class: SPAM
|--- ''m' (+PCA) >  -0.05
|   |--- 'r' (+PCA) <= -0.02
|   |   |--- 't &' (+PCA) <= -0.01
|   |   |   |--- class: HAM
|   |   |--- 't &' (+PCA) >  -0.01
|   |   |   |--- class: SPAM
|   |--- 'r' (+PCA) >  -0.02
|   |   |--- 'free' (+PCA) <= 0.49
|   |   |   |--- class: HAM
|   |   |--- 'free' (+PCA) >  0.49
|   |   |   |--- class: SPAM


|--- 'd' (+PCA) <= -0.02
|   |--- 'un-redeemed' (+PCA) <= -0.00
|   |   |--- class: HAM
|   |--- 'un-redeemed' (+PCA) >  -0.00
|   |   |--- ', love' (+PCA) <= -0.01
|   |   |   |--- class: SPAM
|   |   |--- ', love' (+PCA) >  -0.01
|   |   |   |--- class: HAM
|--- 'd' (+PCA) >  -0.02
|   |--- 'video' (+PCA) <= 0.50
|   |   |--- 'stop' (+PCA) <= 0.96
|   |   |   |--- class: HAM
|   |   |--- 'stop' (+PCA) >  0.96
|   |   |   |--- class: SPAM
|   |--- 'video' (+PCA) >  0.50
|   |   |--- class: SPAM

\end{verbatim}}

\section{Appendix C: Teacher and Student Performance}

As described in \Cref{sec:results}, one can use either the teacher or student model for ARI inference. \Cref{tab:teacherStudent} has the results and suggests that their performance is similar, and that there is no clear winner.

% \usepackage{booktabs}
\begin{table}[H]
\begin{tabular}{@{}lllll@{}}
\toprule
        & SemEval & SMS   & AGNews & CDR   \\ \midrule
Teacher & 65.56   & 95.74 & 91.22  & 57.10 \\
Student & 65.47   & 95.39 & 91.37  & 57.92 \\ \bottomrule
\end{tabular}
\caption{Relative performance of the teacher and student model, using the same filter settings as in \Cref{table:performance_main}. \label{tab:teacherStudent}}
\end{table}

\section{Appendix D: Rule Performance}

\Cref{tab:rulePerformance} gives the performance of the rules by themselves, using the best combination of filters for downstream performance (described in \Cref{sec:results}). Interestingly, we find that the rules do not always outperform BERT, even on the small number of examples they fire on. We hypothesize that the contextualized nature of the teacher's embedding mechanism may be helping it further determine when rules should be applied.

\begin{table}[H]
\begin{tabular}{lllll}
\hline
               & \multicolumn{2}{l}{SemEval} & \multicolumn{2}{l}{AGNews} \\ 
               & Cov.         & Pre.           & Cov.        & Pre.           \\\hline
BERT           & 1.0          & 0.75        & 1.0         & 90.69        \\
Ngram + linear & 0.10             &    47.80   &  0.03   &     96.84         \\
Ngram + tree   &     0.12         &  59.25    &   0.05          &   94.12           \\
PCA + tree     &    0.16          &      50.93     &       0.06      &  94.74            \\ \hline
\end{tabular}
\caption{Performance of BERT and the rules themselves, given as F1 score on the examples where a rule fired. We also provide the coverage, i.e. the proportion of test examples where rules were firing. \label{tab:rulePerformance}}
\end{table}

\section{Appendix E: Samples}

We provide some examples of unigram-based ARI and BERT outputs on the SemEval dataset below. For ease of understanding, we only select examples where only a small number of ngram rules fired.

\begin{enumerate}
\item \textbf{TEXT}: A hinge assembly attaches a cover pivotally to a base of an electronic device and has a pivoting leaf and a stationary leaf . Entity 1: assembly, entity 2: cover. \\ \\
    \textbf{BERT}: Instrument-Agency \\ \\
    \textbf{ARI}: Component-Whole \\ \\
    \textbf{LABEL}: Component-Whole\\ \\
    \textbf{Attn:} \\ \\
    0.60 {\scriptsize\begin{verbatim}
if `has' in x predict Cause-Effect else ABSTAIN
    \end{verbatim}}
    0.97 {\scriptsize\begin{verbatim}
if `has' in x predict Component-Whole else ABSTAIN
    \end{verbatim}}
Interestingly, in this case the same token was mapped to two rules, and the system learned to dynamically prefer one over the other based on context.

\item  \textbf{TEXT}: She left the engine running because the car was full of snakes used in her exotic routine . Entity 1: snakes, entity 2: car. \\ \\
    \textbf{BERT}: Member-Collection \\ \\
    \textbf{ARI}: Content-Container \\ \\
    \textbf{LABEL}: Content-Container \\ \\
    \textbf{Attn:} \\ \\
    0.52 {\scriptsize\begin{verbatim}
if `wa` in x predict CONTENT-CONTAINER else ABSTAIN
    \end{verbatim}}
This is an example of a rule helping the model correctly fix its prediction. ``wa'' often maps to ``was'' with our tokenizer. This rule and the above ``has'' rule are both words that convey a sense of two properties or entities being related to one another, which intuitively seem related to solving the SemEval task (relation classification). 

\item  \textbf{TEXT}: I still shiver as I remember trying to page through economics texts by the flicker from candles while clad in overcoat , scarf , and little knitted gloves with the fingertips cut off , in the 4 p.m . Entity 1: candles, entity 2: flicker. \\ \\
    \textbf{BERT}: Member-Collection \\ \\
    \textbf{ARI}: Cause-Effect \\ \\
    \textbf{LABEL}: Cause-Effect \\ \\
    \textbf{Attn:} \\ \\
    0.78 {\scriptsize\begin{verbatim}
if `,` in x predict MEMBER-COLLECTION else ABSTAIN
    \end{verbatim}}
Interestingly, in this case the rule was incorrect and had high attention but the teacher model (correctly) favored of the student's prediction. Note also that this is a pretty bad rule, as it is a general and nonspecific punctuation marker.

\item  \textbf{TEXT}:  Hands wield the sword in the realm of the flesh , but the intellect wields the pen in the realm of understanding , or of the spirit . Entity 1: pen, entity 2: intellect. \\ \\
    \textbf{BERT}: Instrument-Agency \\ \\
    \textbf{ARI}: Member-Collection \\ \\
    \textbf{LABEL}: Instrument-Agency \\ \\
    \textbf{Attn:} \\ \\
    0.99 {\scriptsize\begin{verbatim}
if `,` in x predict MEMBER-COLLECTION else ABSTAIN
    \end{verbatim}}
This is an example of the same spurious rule as before likely causing the ARI system to make an error. 
\end{enumerate}

% \usepackage{booktabs}
% \begin{table*}[]
% \begin{tabular}{@{}llccccccccc@{}}
% \toprule
% Methods    & AGNews & CDR   & ChemProt & IMDB  & SciCite & SemEval & SMS   & TREC  & Youtube & Avg.  \\ \midrule
% PET        & 91.46  & 51.16 & 53.9     & 88.49 & 75.13   & 64.45   & 95.68 & 83.42 & 95.36   & 75.76 \\
% T-few (3B) & 95.12  & 57.23 & 55.84    & 94.48 & 84.21   & 64.9    & 96.73 & 89.13 & 96.84   & 79.79 \\
% ARI (ours) & 90.87  & 57.92 & 60.01    & 88.33 & 83.76   & 65.1    & 95.74 & 87.19 & 94.95   & 79.05 \\ \bottomrule
% \end{tabular}
% \caption{Comparison between prompt methods PET and T-Few against ARI (PCA+tree variant)}
% \end{table*}

% \usepackage{booktabs}
\begin{table*}[] \small
\begin{tabular}{@{}lllllllllll@{}}
\toprule
                & AGNews & CDR   & ChemProt & IMDB  & SciCite & SemEval & SMS   & TREC  & Youtube & Avg.  \\ \midrule
ARI + BERT            & 90.87  & 57.92 & 60.01    & 88.33 & 83.76   & 65.1    & 95.74 & 87.19 & 94.95   & 79.05   \\
ARI + DeBERTaV3 & 91.38  & 60.09 & 60.48    & 87.19 & 81.97   & 66.41   & 97.20 & 86.28 & 94.88   & 79.40   \\
PET + BERT            & 91.46  & 51.16 & 53.9     & 88.49 & 75.13   & 64.45   & 95.68 & 83.42 & 95.36   & 75.76   \\ 
PET + DeBERTaV3 & 91.14  & 52.63 & 54.84    & 88.92 & 75.97   & 63.50   & 96.14 & 84.90 & 95.09   & 76.29  
\end{tabular}
\caption{Comparison between PET and ARI (PCA+tree variant) with BERT and DeBERTaV3 backbones. Both methods improved slightly (PET by 0.65\% and ARI by 0.11\% on average) but ARI remains better overall.\label{table:encoders}}
\end{table*}

% \usepackage{booktabs}
\begin{table}[]
\small
\centering
\begin{tabular}{@{}lllllllllll@{}}
\toprule
            & AGNews & CDR   & ChemProt & IMDB  & SciCite & SemEval & SMS   & TREC  & Youtube & Avg. \\ \midrule
BERT       & 90.33  & 69.56 & 64.12    & 86.84 & 79.84   & 82.23   & 98.6  & 86.8 & 91.66   & 82.66   \\
SELF       & 89.52  & 70.65 & 65.34    & 86.92 & 80.21   & 85.43   & 98.6  & 87.6 & 93.33   & 83.55   \\
VAT        & 89.51  & 72.28 & 67.98    & 89.51 & 80.28   & 87.64   & 99.2  & 87.4 & 94.16   & 84.77   \\ 
PET        & 90.53  & 69.48 & 65.84    & 88.71 & 78.12   & 87.5    & 96.73 & 86.8 & 96.83   & 83.81   \\
PR-BOOST   &   88.9     &       & 67.1     &       &         &         &       &      &         &         \\
ARI (ours) & 89.76  & 72.06 & 68.71    & 89.68 & 81.67   & 87.08   & 99    & 87.4 & 97.5    & 85.30   \\
T-few (3B)    & 94.2   & 72    & 70.98    & 94.46 & 82.9    & 86.9    & 99.15 & 95   & 95      & 87.2    \\ \bottomrule
\end{tabular}
\centering
\caption{Accuracies for the main results (Table 2).\label{tab:acc}}
\end{table}

\end{document}